\documentclass{article}

\usepackage{microtype}
\usepackage{graphicx}
\usepackage{subcaption}
\usepackage{booktabs}   
\usepackage{colortbl}   
\usepackage{xcolor}     
\usepackage{tabularx}          
\usepackage{array}             
\usepackage[most]{tcolorbox} 
\usepackage{setspace}

\newcommand{\codeindent}{\hspace*{2em}}
\newcommand{\codeindenttwo}{\hspace*{4em}}
\newcommand{\codeindentthree}{\hspace*{6em}}

\usepackage{hyperref}

\usepackage[preprint]{icml2026}

\usepackage{amsmath}

\usepackage{amssymb}
\usepackage{mathtools}
\usepackage{amsthm}

\usepackage[capitalize,noabbrev]{cleveref}

\theoremstyle{plain}

\theoremstyle{definition}

\theoremstyle{remark}

\usepackage[textsize=tiny]{todonotes}

\icmltitlerunning{FunPRM}

\begin{document}

\twocolumn[
  \icmltitle{FunPRM: Function-as-Step Process Reward Model with Meta Reward Correction for Code Generation}

  \icmlsetsymbol{equal}{*}

  \begin{icmlauthorlist}
    \icmlauthor{Ruiyi Zhang}{ucsd}
    \icmlauthor{Peijia Qin}{ucsd}
    \icmlauthor{Qi Cao}{ucsd}
    \icmlauthor{Eric Xue}{ucsd}
    \icmlauthor{Pengtao Xie}{ucsd}
  \end{icmlauthorlist}

  \icmlaffiliation{ucsd}{University of California, San Diego}
  \icmlcorrespondingauthor{Pengtao Xie}{p1xie@ucsd.edu}

  \icmlkeywords{Machine Learning, ICML}

  \vskip 0.3in
  
]

\printAffiliationsAndNotice{}  

\begin{abstract}
Code generation is a core application of large language models (LLMs), yet LLMs still frequently fail on complex programming tasks. Given its success in mathematical reasoning, test-time scaling approaches such as Process Reward Model (PRM)-based Best-of-N selection offer a promising way to improve performance. However, existing PRMs remain ineffective for code generation due to the lack of meaningful step decomposition in code and the noise of Monte Carlo-estimated partial-solution correctness scores (rewards). To address these challenges, we propose FunPRM. FunPRM prompts LLMs to encourage modular code generation organized into functions, with functions treated as PRM reasoning steps. Furthermore, FunPRM introduces a novel meta-learning-based reward correction mechanism that leverages clean final-solution rewards obtained via a unit-test-based evaluation system to purify noisy partial-solution rewards. Experiments on LiveCodeBench and BigCodeBench demonstrate that FunPRM consistently outperforms existing test-time scaling methods across five base LLMs, notably achieving state-of-the-art performance on LiveCodeBench when combined with O4-mini. Furthermore, FunPRM produces code that is more readable and reusable for developers. 

\end{abstract}

\section{Introduction }

Code generation has become one of the most widely used and economically significant applications of large language models (LLMs)~\citep{huang2025aiwork,anthropic2026aeiv4}. However, even state-of-the-art LLMs frequently hallucinate and make observable errors~\citep{jiang2025survey}, largely due to the complexity of multi-step reasoning required for non-trivial programming tasks~\citep{yu2025reasoning}. Since LLMs often produce both correct and incorrect solutions when sampled multiple times, effective Best-of-$N$ solution selection strategies can substantially improve performance~\citep{wang-etal-2024-multi-step}. Among these approaches, Process Reward Model (PRM)-based selection has gained popularity~\citep{lightman2024lets}. Rather than evaluating only the final solution, PRMs assess whether intermediate reasoning steps of LLMs make progress toward a correct solution. This step-level evaluation enables PRMs to more precisely discriminate between promising and flawed solutions, making them well suited for Best-of-$N$ selection. PRMs have demonstrated strong effectiveness in reasoning tasks, such as mathematical problem solving~\citep{lightman2024lets} and multimodal question answering~\citep{cao2025dreamprm}, making them a promising direction for enhancing LLM-based code generation.

Despite this potential, two challenges prevent existing PRMs from being effectively used for coding. First, most PRMs require LLM-generated solutions to be separated into meaningful reasoning steps. Unlike LLM solutions to mathematical problems, which naturally decompose into explicit reasoning steps under Chain-of-Thought (CoT) prompting~\citep{wei2022chain}, LLM-generated code lacks an obvious definition of such ``steps.'' Prior work often treats each line of code as a step, which can lead to hundreds of steps for certain programs, significantly increasing computation cost.~\citep{wang-etal-2024-multi-step, He2025SkyworkOR}. Second, it is difficult to efficiently obtain the ground-truth correctness scores for partial solutions, which serve as labels to train PRMs. Some works use human-labeled scores, which are costly to obtain at large scale~\citep{lightman2024lets}. Others rely on Monte Carlo-based methods to automatically obtain such scores, typically by estimating the likelihood that a partial solution leads to a correct final solution via sampling~\citep{wang-etal-2024-math,Luo2024ImproveMR}. Despite the reduced cost, scores obtained from Monte Carlo sampling can be quite noisy and ultimately reduce PRM performance.

\begin{figure*}[t]
    \centering
    \includegraphics[width=0.98\linewidth]{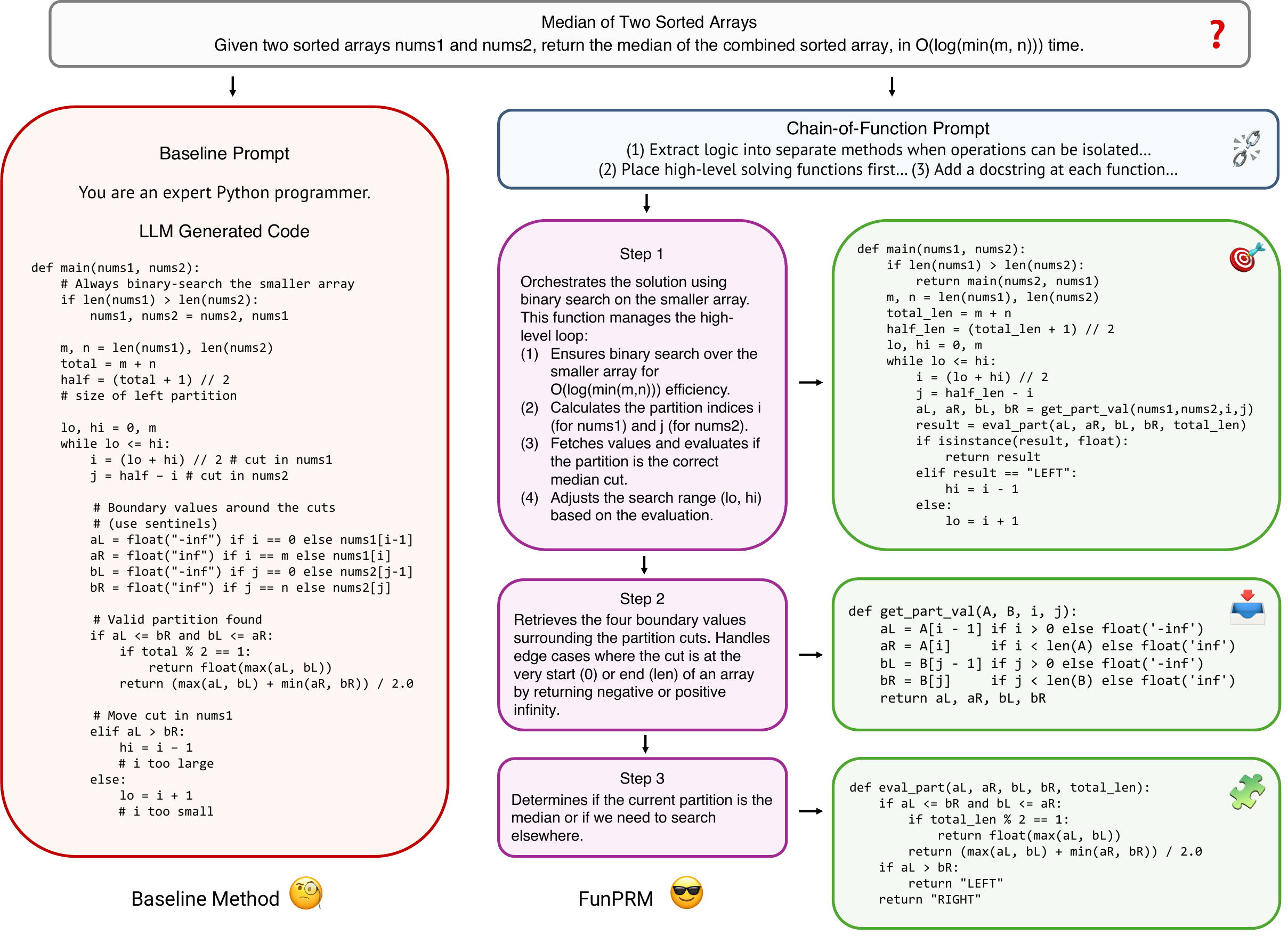}
    \caption{\textbf{Comparison of generated code between the baseline method and FunPRM.} 
    FunPRM prompts LLMs to encourage them generate modular code organized into multiple functions with accompanying docstrings. These functions serve as reasoning steps for the process reward model while simultaneously improving code readability for human developers.}
    \label{fig:CoF}
\end{figure*}

To address these challenges, we propose FunPRM, a process reward model tailored to code generation.
Firstly, we prompt LLMs to generate modular code that organizes logically independent operations into separate functions, and treat each function as a ``step'' for PRM. As illustrated in Figure~\ref{fig:CoF} with an example of computing the median of two arrays, the first step is the main function, which outlines a high-level strategy based on binary search. The second step implements a helper function to compute boundary values for a partition, and the third step checks whether the current partition is the median.  
Secondly, FunPRM introduces a meta-learning-based reward correction mechanism. We observe that, although Monte Carlo-estimated correctness scores (rewards) for partial solutions are noisy, the correctness of final solutions to a coding problem can be reliably determined by a unit-test-based evaluation system. Leveraging this property, we propose a bi-level meta-learning scheme that uses clean final-solution rewards to purify noisy rewards for partial solutions. Specifically, we first train the PRM using Monte Carlo-estimated rewards for partial solutions and perform a one-step gradient update. We then evaluate the updated PRM on final solutions and use the resulting loss to compute gradients with respect to the noisy partial-solution rewards, which are subsequently optimized to improve their quality. 

Our key contributions are as follows:
\begin{itemize}
    \item We propose FunPRM, a process reward model tailored to code generation. FunPRM prompts LLMs to encourage the use of functions in generated code and treats functions as reasoning steps for PRM evaluation. In addition, leveraging clean final-solution rewards, FunPRM introduces a meta-learning-based reward correction mechanism to denoise Monte Carlo-sampled partial-solution rewards used for PRM training.
    \item We evaluate FunPRM on two large-scale code generation benchmarks under Best-of-$N$ selection, where it consistently outperforms a wide range of test-time scaling baselines in terms of pass@1 across five base LLMs. Notably, when combined with OpenAI O4-mini (High), FunPRM achieves state-of-the-art performance on LiveCodeBench with 80.9 pass@1. Human evaluations further show that FunPRM-generated code is preferred by developers in terms of readability and reusability.
\end{itemize}

\begin{figure}[t]
    \centering
    \includegraphics[width=\linewidth]{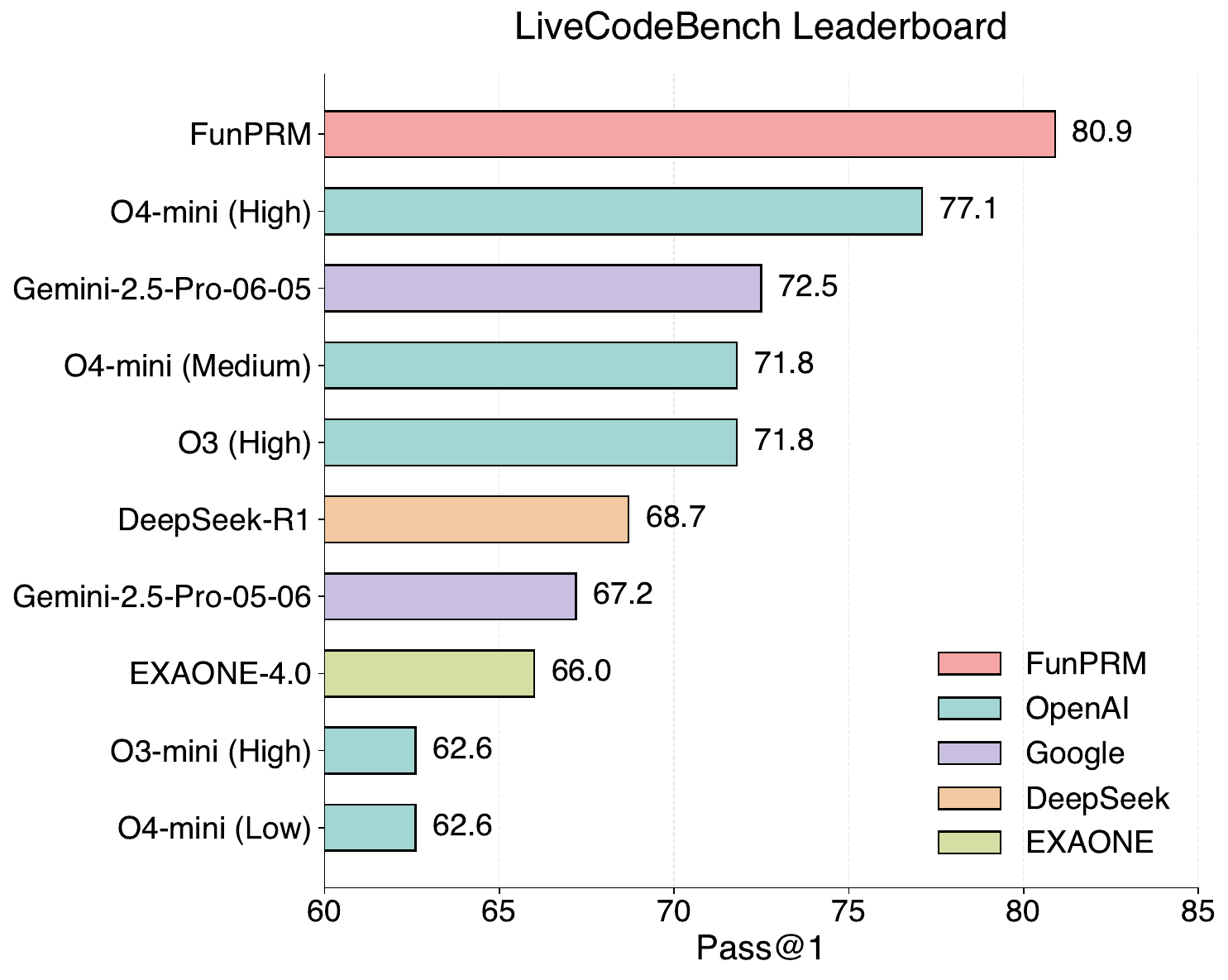}
    \caption{\textbf{Leaderboard results on LiveCodeBench (2025-02-01--present) for FunPRM and other LLMs.} FunPRM achieves state-of-the-art performance when using OpenAI O4-mini (High) as the base LLM.}
    \label{fig:leaderboard}
\end{figure}

\section{Related Works}

\paragraph{Process Reward Models}
Process Reward Models (PRMs) have been widely studied for improving the reasoning capabilities of large language models (LLMs), particularly in mathematical domains. \citet{lightman2024lets} is among the first to introduce PRMs trained using human-annotated step-wise reward data. To reduce the cost of human annotation, MiPS~\citep{wang-etal-2024-multi-step} proposes a Monte Carlo (MC) sampling-based approach for automatic stepwise reward labeling, enabling performance gains through best-of-$N$ selection. OmegaPRM~\citep{Luo2024ImproveMR} introduces an MCTS-based automatic labeling strategy and improves base LLM performance via reinforcement learning. ReST-MCTS~\citep{zhang2024restmcts} further integrates PRM labeling with reinforcement learning of the base LLM, forming a self-training framework. DreamPRM addresses data quality issues in PRM training through a meta-learning-based domain reweighting approach~\citep{cao2025dreamprm}.

\paragraph{Test-Time Approaches for Code Generation}
Due to the free-form nature of code generation, some test-time scaling (TTS) methods, such as majority voting, are not directly applicable to coding tasks~\citep{wang2023selfconsistency}. Self-Certainty proposes using the model’s confidence over generated code as a criterion for best-of-$N$ selection~\citep{kang2025scalable}. MiPS and Skywork-PRM extend PRM-based test-time scaling to coding by adopting strategies similar to those used in mathematical reasoning, treating each line of code as a reasoning step~\citep{He2025SkyworkOR,wang-etal-2024-multi-step}. Recently, LLM-as-a-Judge has been applied to test-time scaling of LLMs~\citep{zhou2025evaluating}, including applications to code generation~\citep{QinDAJ2026}, often incurring inference cost that scales quadratically with the number of candidate solutions due to pairwise comparisons. In parallel, execution-feedback-enhanced agentic TTS frameworks have gained popularity in code generation, leveraging multi-round generation and repeated code execution with additional public test cases to guide iterative refinement. Reflexion converts execution feedback into textual guidance for iterative code refinement~\citep{shinn2023reflexion}, while LDB collects step-by-step execution feedback to provide fine-grained supervision signals~\citep{zhong-etal-2024-debug}. ORPS treats each round of code generation and execution as a reasoning step for PRM-guided decoding~\citep{yu2025reasoning}. CodePRM introduces an explicit planning stage prior to code generation, treating natural-language plans as PRM steps and incorporating code execution feedback into the reward model input~\citep{li-etal-2025-codeprm}. Despite its name, CodePRM does not conform to the standard definition of a process reward model, as it cannot directly score partial solutions, since code execution feedback is unavailable for partial programs.

\paragraph{Meta Label Correction}
Meta-learning has been widely adopted for correcting corrupted labels, particularly in computer vision. M-SLC is among the first methods to formulate label correction as a meta-learning problem, learning a trainable label corrector using a small set of clean labels~\citep{Wu2020LearningTP}. EMLC extends this line of work by introducing a novel meta-gradient approximation and a teacher model to further improve label correction performance~\citep{Taraday_2023_ICCV}. DMLP further integrates a label-free representation learning phase into the meta-learning framework, enhancing robustness to noisy supervision~\citep{Tu2023LearningFN}.  More recently, RobPicker introduces a unified framework for label correction and data reweighting in segmentation tasks~\citep{Hosseini2025.09.16.676650}. However, to the best of our knowledge, no prior work has explored label correction methods for improving the training and effectiveness of process reward models.
\section{Preliminaries}

\paragraph{Definitions}
We first introduce the standard formulation of Process Reward Models (PRMs) for test-time scaling in multi-step reasoning tasks~\citep{wang-etal-2024-multi-step}. Given a reasoning dataset $\mathcal{D}=\{(x,y)\}$, a base LLM (policy model) takes the input $x$ as the initial state $s_0$ and iteratively generates a reasoning trajectory $\tau = (s_0, s_1, \ldots, s_T)$ according to its policy $\pi_\theta(s_{t+1} \mid s_t)$. Here, $s_T$ corresponds to the final solution generated by the model, while each intermediate state $s_t$ represents a partial solution that is a prefix of the final solution, as commonly produced under Chain-of-Thought (CoT) prompting.
A PRM $f_\phi$ assigns a scalar correctness score (reward) to each partial solution, $r_t = f_\phi(s_t)$, reflecting the likelihood that the reasoning trajectory up to step $t$ will lead to a correct final answer. Given a trained PRM, Best-of-$N$ test-time scaling selects the solution with the highest aggregated reward, typically computed by averaging rewards across steps:
\begin{equation}
\label{eq:prm-score}
    f_\phi(\tau) = \frac{1}{T}\sum_{t=1}^{T} f_\phi(s_t).
\end{equation}

\paragraph{Training of Process Reward Models}

A central challenge in training Process Reward Models (PRMs) is the absence of ground-truth correctness scores $r_t$ for partial solutions. Early work addresses this issue by relying on human annotators to manually label these scores~\citep{lightman2024lets}, but this approach is prohibitively expensive at scale. As a result, Monte Carlo (MC)-based strategies that automatically estimate partial-solution rewards have become popular~\citep{Luo2024ImproveMR,cao2025dreamprm}.  
Given a partial solution $s_t$, the MC estimator uses the base LLM to sample $K$ complete solutions:
\begin{equation}
    s_T^{(k)} \sim \pi_\theta(\cdot \mid s_t), \quad k = 1, \ldots, K.
\end{equation}
It then computes the fraction $r_t'$ of these $K$ solutions whose final predicted labels match the ground-truth label $y$, and uses this value as a Monte Carlo estimate of the partial-solution reward $r_t$. The PRM, typically implemented as a text classification model, is trained to regress these estimated rewards using a mean squared error loss.

\section{Method} 

In this section, we describe the proposed FunPRM framework in detail. FunPRM is a Process Reward Model tailored for code generation, which defines reasoning steps at the level of functions and incorporates a reward correction mechanism to improve the quality of training signals.

\subsection{Functions in Code as PRM Steps}
To address the first challenge discussed above—the lack of a natural step decomposition in code—we propose a Chain-of-Function prompting strategy. This strategy encourages LLMs to generate modular code organized into functions, allowing each function to serve as a reasoning step for the PRM. Concretely, the prompt guides the LLM to group logically independent code blocks into separate functions, with higher-level functions (e.g., the \texttt{main} function) appearing first. In addition, the prompt encourages the model to write docstrings at the beginning of each function, which act as high-level specifications for the corresponding implementations.

The prompt template is shown in Figure~\ref{fig:CoF-prompt}, and an example of the resulting code is illustrated in Figure~\ref{fig:CoF}. In the first step, the \texttt{main} function outlines the overall strategy by using a binary search partition to locate the correct split between two sorted arrays. The second step, \texttt{get\_part\_val}, computes the four boundary elements around a candidate partition, using sentinel values to handle edge cases. The third step, \texttt{eval\_part}, checks whether the partition satisfies the ordering invariant and determines whether the search should proceed to the left or right. By defining reasoning steps at the function level, this formulation yields a clear and semantically meaningful notion of PRM steps, enabling effective PRM training and inference while producing modular code with improved readability and reusability.

\newtcolorbox{datacollection}[1]{%
  enhanced,
  colback=gray!10,
  colframe=gray!100,
  boxrule=1.5pt,
  leftrule=1.5pt,
  rightrule=1.5pt,
  toprule=1.5pt,
  bottomrule=1.5pt,
  left=3.5mm,
  right=3.5mm,
  top=3.5mm,
  bottom=3.5mm,
  width=\columnwidth,
  before upper={\setlength{\parskip}{0pt}\setlength{\parindent}{0pt}},
  before skip=8pt,
  after skip=8pt,
}

\begin{figure}[t]
  \centering
  \begin{minipage}{0.95\linewidth}
\begin{datacollection}

{You will be given a question (problem specification) and will generate a correct Python program that matches the specification and passes all tests. Follow the code organization guidelines below:}

\vspace{0.6em}
\noindent\textbf{Logic decomposition}:
{Extract logic into separate functions when operations are repeated or when complex calculations can be isolated.}

\vspace{0.6em}
\noindent\textbf{Function organization}: {Place high-level solving functions first (e.g., {main()}). Follow with helper functions that implement specific subtasks. Avoid nested functions; keep all functions at the top level.}

\vspace{0.6em}
\noindent\textbf{Docstrings in each function}: {Add a docstring using triple quotes at the beginning of each function. Main methods should explain the approach, algorithm choice, strategy, and key steps; helper methods should explain their specific logic and purpose.}

\vspace{0.6em}
\textbf{Example demonstration}: The following example illustrates the code structure and docstring style only. Solve the actual problem, \emph{not} the example algorithm.

\vspace{0.6em}
[Few-shot examples]

\end{datacollection}
  \end{minipage}
  \caption{\textbf{Chain-of-Function system prompt.} The prompt encourages function-level logic decomposition, top-down function organization, and descriptive docstrings to produce clearly defined PRM reasoning steps in generated code.}
  \label{fig:CoF-prompt}
\end{figure}


\subsection{Automatic Reward Correction with Meta Learning}

\begin{figure}[t]
    \centering
    \includegraphics[width=\linewidth]{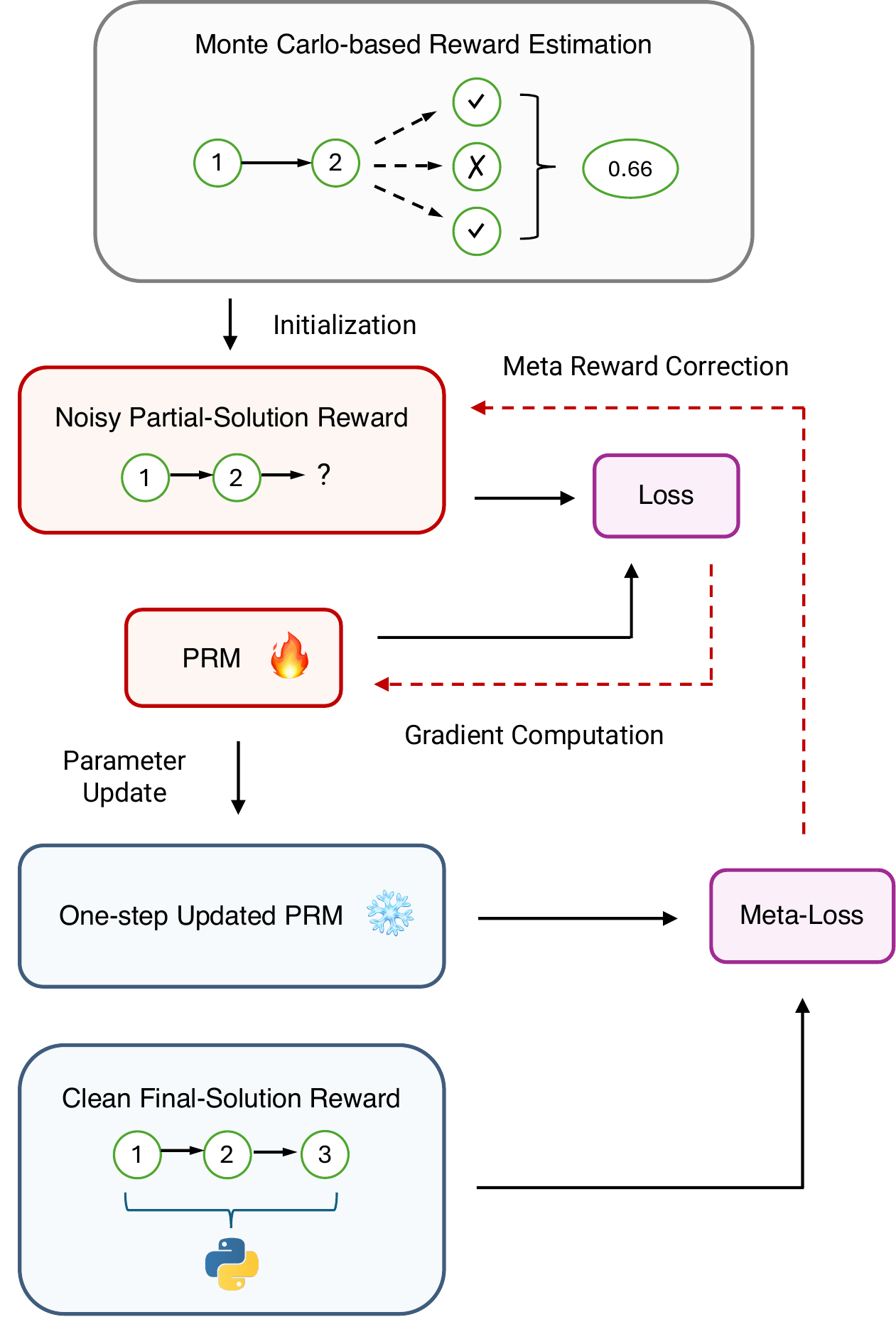}
    \caption{\textbf{Meta-learning-based reward correction framework in FunPRM.}
Noisy partial-solution rewards (correctness scores) are initialized via Monte Carlo sampling. The PRM parameters are first updated using these noisy partial-solution rewards through a one-step gradient descent. The updated PRM is then evaluated on clean final-solution reward data to compute a meta loss, which is used to optimize the partial-solution rewards from the previous stage.}
    \label{fig:method}
\end{figure}

To improve the quality of Monte Carlo-sampled PRM training data, we propose a meta-learning-based reward correction framework that explicitly denoises correctness scores (rewards) for partial solutions. In coding tasks, we can leverage a private unit-test evaluation system $E$ to obtain clean and reliable final-solution rewards $r_T = E(s_T)$ once the LLM completes code generation and produces a final solution. These rewards provide high-quality supervision signals that are generally unavailable in mathematical reasoning tasks, where a solution $s_T$ is typically deemed correct if it contains the ground-truth answer $y$. This assumption, however, is imperfect, as incorrect reasoning steps may still occasionally lead to a correct final answer~\citep{turpin2023language}. FunPRM exploits this distinctive property of coding problems to improve the quality of partial-solution rewards and, in turn, the effectiveness of PRM training.

For each trajectory $\tau = (s_1, s_2, \ldots, s_T)$ in the set of trajectories $\mathcal{T}$ generated by an LLM as candidate solutions to a coding dataset, we consider two sources of supervision. The first is a clean meta-dataset of final solutions $(S_m, R) = \{(s_T(\tau), r_T(\tau)) \mid \tau \in \mathcal{T}\}$, where final-solution rewards are obtained from a unit-test evaluation system. The second is a noisy dataset of partial solutions $(S_n, \widehat{R}) = \{(s_t(\tau), \hat{r}_t(\tau)) \mid \tau \in \mathcal{T},\; t < T(\tau)\}$, where partial-solution rewards (correctness scores) are estimated via Monte Carlo-sampling.
Since Monte Carlo-estimated partial rewards are noisy, we introduce a lightweight, trainable reward-correction table $g_\theta$ to refine $\widehat{R}$. The correction table adds a learnable residual to each partial-solution reward, followed by clamping the corrected value to the range $[0,1]$. We adopt a reward-correction table rather than a parametric network because coding datasets are typically small, making overparameterized models prone to overfitting. This design enables adaptive correction of noisy partial rewards guided by PRM performance on clean final-solution rewards.

The proposed reward-correction table is optimized using a meta-learning framework. First, the PRM $f_{\phi}$ is trained on the partial-solution dataset $S_n$ using the current corrected rewards $g_\theta(\widehat{R})$. The PRM parameters $\phi$ are updated to $\hat{\phi}$ by minimizing the training loss with a single gradient descent step:
\begin{equation}
\label{eq:lower}
    \hat{\phi} = \phi - \eta \nabla_{\phi}\mathcal{L}\Big(f_\phi(S_n),\, g_\theta(\widehat{R})\Big),
\end{equation}
where $\eta$ denotes the learning rate. Note that the updated parameters $\hat{\phi}$ depend implicitly on the reward-correction parameters $\theta$ through the corrected partial-solution rewards $g_\theta(\widehat{R})$.
Given this one-step update, we next evaluate the updated PRM $f_{\hat{\phi}}$ on the clean meta-dataset $(S_m, R)$ of final solutions and define the following meta-objective over the reward-correction parameters $\theta$:
\begin{equation}
\label{eq:meta}
    \min_{\theta}\; \mathcal{L}\Big(f_{\hat{\phi}}(S_m),\, R\Big).
\end{equation}
Because the meta-loss depends on $\hat{\phi}$ and, through the inner update, implicitly on $\theta$, minimizing this objective encourages corrected partial-solution rewards that lead to PRM parameters generalizing well to clean final-solution rewards.
In practice, computing the gradient of Eq.~\ref{eq:meta} with respect to $\theta$ involves second-order derivatives. To reduce computational overhead, we adopt a finite-difference approximation to estimate the meta-gradient~\citep{liu2018darts,choe2023betty}, with details provided in Appendix~\ref{appen:meta-grad}. Through this meta-learning procedure, FunPRM progressively refines noisy partial-solution rewards, yielding denoised training signals that improve PRM robustness and downstream performance in test-time scaling. An overview of the reward correction process is illustrated in Figure~\ref{fig:method}.

\section{Results}

\subsection{Experimental Settings}

\paragraph{Datasets} We evaluate FunPRM primarily on LiveCodeBench (LCB)~\citep{jain2025livecodebench} and BigCodeBench (BCB)~\citep{zhuo2025bigcodebench}. We use two other programming datasets, HumanEval+~\citep{Chen2021EvaluatingLL} and MBPP+~\citep{austin2021program} for domain generalization experiments, evaluated by EvalPlus  system~\citep{liu2023is}.  All settings used in this work ensure no overlap between training and evaluation data. More detailed descriptions of these datasets and their usage are provided in Appendix~\ref{appen:data}.

\paragraph{Training Settings}  We adopt Qwen-2.5-Coder-7B~\citep{Yang2024Qwen25TR} as the backbone for FunPRM under a generative process reward model setting~\citep{zhang2025generative}, as detailed in Appendix~\ref{appen:gprm}. During training, only the LoRA layers of the PRM are updated~\citep{hu2022lora}. Additional details on generating PRM training data, PRM training settings and hyperparameters are provided in Appendix~\ref{appen:hyperparam}.

\paragraph{Test-time Settings}For test-time scaling, we use OpenAI O4-mini~\citep{openai_o3_o4mini_system_card} with high reasoning effort, Qwen3-Coder-30B-A3B~\citep{Yang2025Qwen3TR}, GPT-4o-mini~\citep{Hurst2024GPT4oSC}, DeepSeek-Coder~\citep{Guo2024DeepSeekCoderWT}, and  Qwen2.5-7B-Coder~\citep{Yang2024Qwen25TR} as the base (policy) model. For each problem, the base LLM generates eight candidate solutions, from which FunPRM selects the final output. 

\begin{table}[t]
\centering
\setlength{\tabcolsep}{5pt}
\caption{\textbf{Performance comparison of FunPRM and baseline methods across Easy, Medium, and Hard difficulty levels on LiveCodeBench (2025-02-01 -- present).} Results include top-performing LLMs from the leaderboard and test-time scaling methods applied to O4-mini (High), with results reported as pass@1 (\%). Bold values indicate the best performance in each column.}
\begin{tabular}{lcccc}
\toprule
{\textbf{Model}} &
{\textbf{Easy}} &
{\textbf{Medium}} &
{\textbf{Hard}} &
{\textbf{Overall}} \\
&(31)&(39)&(61)&(131)\\

\midrule

\rowcolor{gray!15}
\multicolumn{5}{c}{{Top LLMs on LiveCodeBench Leaderboard}}\vspace{0.1cm} \\
Gemini-2.5      & \textbf{100} & 82.1 & 52.5 & 72.5 \\
O3 (High)             & \textbf{100} & 71.8 & 57.4 & 71.8 \\
DeepSeek-R1     & 99.7 & 77.7 & 47.2 & 68.7 \\
O4-mini (High)    & \textbf{100} & 89.7 & 57.4 & 77.1 \\

\midrule
\rowcolor{gray!15}
\multicolumn{5}{c}{{Test-time Scaling Methods (O4-mini)}}\vspace{0.1cm} \\
Self-Certainty  & \textbf{100} & 86.5 & 59.8 & 77.3 \\
ORM             & \textbf{100} & 89.7 & 62.3 & 79.4 \\
Skywork-PRM     & \textbf{100} & 87.2 & 59.8 & 78.1 \\
\textbf{FunPRM} & \textbf{100} & \textbf{92.3} & \textbf{63.9} & \textbf{80.9} \\

\bottomrule
\end{tabular}
\label{tab:leaderboard}
\end{table}

\paragraph{Baselines}
We compare FunPRM against three alternative test-time scaling (TTS) approaches: Self-Certainty~\citep{kang2025scalable}, Outcome Reward Models (ORMs)~\citep{Cobbe2021TrainingVT,Uesato2022SolvingMW}, and Skywork-PRM~\citep{He2025SkyworkOR}. In addition, we report results for several strong LLMs on LiveCodeBench leaderboard without test-time scaling. 
We focus our comparisons on Best-of-N test-time scaling baselines that perform a single round of code generation without using test-case information at inference time. In contrast, agentic, multi-round, execution-enhanced approaches~\citep{li-etal-2025-codeprm,yu2025reasoning,li-etal-2025-test} are not directly comparable for two reasons: (i) they incur substantially higher inference cost due to multi-round generation and repeated code execution, and (ii) they rely on public test-case information that is not consistently available in benchmark problem statements, such as MBPP~\citep{austin2021program} and BigCodeBench~\citep{zhuo2025bigcodebench}. 
We present more details of these baselines in Appendix~\ref{appen:baseline}.

\subsection{FunPRM Achieves Top-1 Performance on LiveCodeBench with a State-of-the-Art Base LLM}

\begin{table*}[t]
\centering
\setlength{\tabcolsep}{12pt}
\caption{\textbf{Performance comparison of test-time scaling methods on LiveCodeBench (2024-08-01 -- present) and BigCodeBench across different base models. }
LiveCodeBench results are reported as pass@1 (\%) on Easy (110), Medium (141), and Hard (203) subsets, together with the Overall score on all 454 problems. 
BigCodeBench results are reported as pass@1 (\%) on Easy (292) and Hard (48) subsets, along with the Overall score on all 340 problems. 
Best results within each base-model block are shown in bold.}
\label{tab:result}
\begin{tabular}{l|cccc|ccc}
\toprule
& \multicolumn{4}{c}{\textbf{LiveCodeBench}} & \multicolumn{3}{c}{\textbf{BigCodeBench}} \\
\cmidrule(lr){2-5} \cmidrule(lr){6-8}
\textbf{Method} &
\textbf{Easy} &
\textbf{Medium} &
\textbf{Hard} &
\textbf{Overall} &
\textbf{Easy} &
\textbf{Hard} &
\textbf{Overall} \\
& (110) & (141) & (203) & (454) & (292) & (48) & (340) \\
\midrule

\rowcolor{gray!15}
\multicolumn{8}{c}{Base Model: Qwen3-30B-A3B}\vspace{0.1cm} \\
Base             & 87.05 & 31.65 & 4.93 & 33.12 & 48.92 & 30.24 & 46.29 \\
Self-Certainty   & 83.64 & 24.82 & 2.46 & 29.07 & 49.74 & 31.25 & 47.06 \\
ORM              & 88.18 & 34.68 & 7.39 & 35.44 & 49.74 & 29.17 & 46.76 \\
Skywork-PRM      & \textbf{89.09} & 32.62 & 4.93 & 33.92 & 48.29 & 31.08 & 45.88 \\
FunPRM           & 88.18 & \textbf{36.17} & \textbf{8.37} & \textbf{36.34} & \textbf{50.51} & \textbf{32.43} & \textbf{47.94} \\

\midrule
\rowcolor{gray!15}
\multicolumn{8}{c}{Base Model: GPT-4o-mini}\vspace{0.1cm} \\
Base             & 78.18 & 18.90 & 3.90 & 27.50 & 44.20 & {27.08} & 41.80 \\
Self-Certainty   & 79.09 & 22.70 & 4.43 & 28.19 & 44.26 & 22.92 & 41.18 \\
ORM              & 78.18 & 23.58 & 4.80 & 28.40 & 43.23 & 25.00 & 40.59 \\
Skywork-PRM      & 80.00 & \textbf{24.82} & 4.93 & 29.30 & 44.56 & {27.08} & 42.06 \\
FunPRM           & \textbf{80.91} & \textbf{24.82} & \textbf{5.42} & \textbf{29.74} & \textbf{45.89} & \textbf{31.25} & \textbf{43.82} \\

\midrule
\rowcolor{gray!15}
\multicolumn{8}{c}{Base Model: DeepSeek-Coder-33B-Instruct}\vspace{0.1cm} \\
Base             & 54.55 & 12.68 & 1.91 & 18.00 & 44.10 & 21.09 & 40.90 \\
Self-Certainty   & 50.00 &  9.93 & 0.99 & 15.64 & 41.03 & 20.00 & 37.93 \\
ORM              & 57.27 & 12.68 & 1.91 & 18.67 & 43.84 & 20.83 & 40.59 \\
Skywork-PRM      & 57.27 & 14.89 & 1.97 & 19.40 & 44.18 & 20.00 & 40.69 \\
FunPRM           & \textbf{63.64} & \textbf{15.60} & \textbf{2.46} & \textbf{21.37} & \textbf{48.06} & \textbf{25.71} & \textbf{44.83} \\

\midrule
\rowcolor{gray!15}
\multicolumn{8}{c}{Base Model: Qwen2.5-7B-Coder}\vspace{0.1cm} \\
Base             & 54.77 &  8.33 & 2.28 & 16.87 & 37.79 & 27.08 & 36.30 \\
Self-Certainty   & 57.27 &  5.67 & 0.99 & 16.08 & 35.28 & 27.08 & 34.10 \\
ORM              & 58.30 &  9.31 & 2.28 & 18.03 & 40.30 & \textbf{31.25} & 39.04 \\
Skywork-PRM      & 59.09 & \textbf{10.64} & 2.46 & 18.72 & 38.68 & 29.17 & 37.35 \\
FunPRM           & \textbf{63.64} &  9.22 & \textbf{2.96} & \textbf{19.60} & \textbf{41.12} & \textbf{31.25} & \textbf{39.71} \\

\bottomrule
\end{tabular}
\end{table*}

The results of FunPRM with OpenAI O4-mini (High) and the baselines are reported in Figure~\ref{fig:leaderboard} and Table~\ref{tab:leaderboard}. FunPRM outperforms all leading proprietary models on the LiveCodeBench leaderboard in terms of pass@1 percentage, validating the effectiveness of test-time scaling (TTS) in substantially pushing the performance frontier even for already high-performing LLMs. We further compare FunPRM with other TTS methods under the same base model, O4-mini (High), where FunPRM consistently achieves the best performance, highlighting the advantages of its test-time scaling strategy. In summary, the state-of-the-art results of FunPRM demonstrate the effectiveness of the proposed approach when combined with strong base LLMs.

\subsection{Benchmarking Evaluation of FunPRM}
We conduct a comprehensive evaluation of FunPRM against multiple test-time scaling baselines across four popular base LLMs on both LiveCodeBench and BigCodeBench. As shown in Table~\ref{tab:result}, FunPRM  outperforms all baseline methods in terms of overall performance across all four base models and both benchmarks. These results demonstrate the robustness of FunPRM when paired with different base LLMs and applied to diverse styles of coding tasks.  
FunPRM surpasses the Self-Certainty baseline, indicating the necessity of a learned reward model compared to simpler heuristic-based test-time scaling strategies. It also outperforms outcome reward model (ORM)-based scaling, highlighting the advantage of decomposing code generation into intermediate steps and assigning rewards at a finer granularity, and underscoring the importance of the proposed Chain-of-Function (CoF) formulation. In addition, FunPRM outperforms Skywork-PRM, demonstrating the effectiveness of the meta-learning-based reward correction mechanism.  
We further report results on individual subsets, where FunPRM achieves the best performance in 18 out of 20 settings, showcasing its strong and consistent capability across coding tasks with varying difficulty levels.

\begin{figure}[h!]
    \centering
    \includegraphics[width=0.8\linewidth]{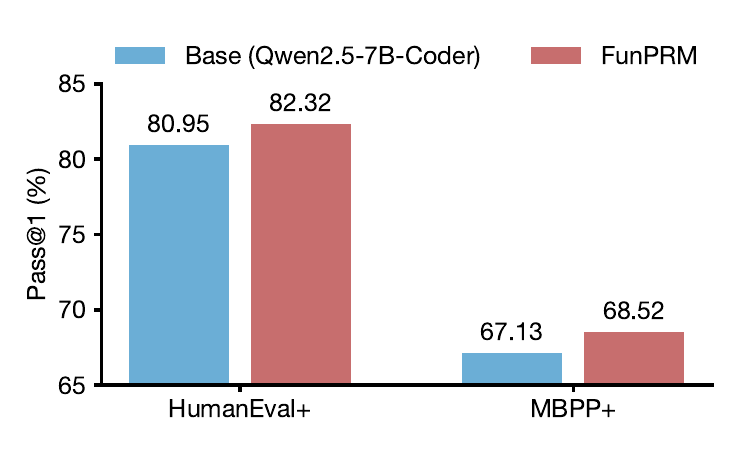}
    \caption{
    \textbf{Domain generalization results of FunPRM on HumanEval+ and MBPP+.}
    FunPRM consistently improves pass@1 over the base Qwen2.5-7B-Coder model,
    indicating improved code generation quality across both benchmarks.
    }
    \label{fig:domain-gen}
\end{figure}

\subsection{Domain Generalization}

\begin{figure*}[t]
    \centering
    \includegraphics[width=0.88\linewidth]{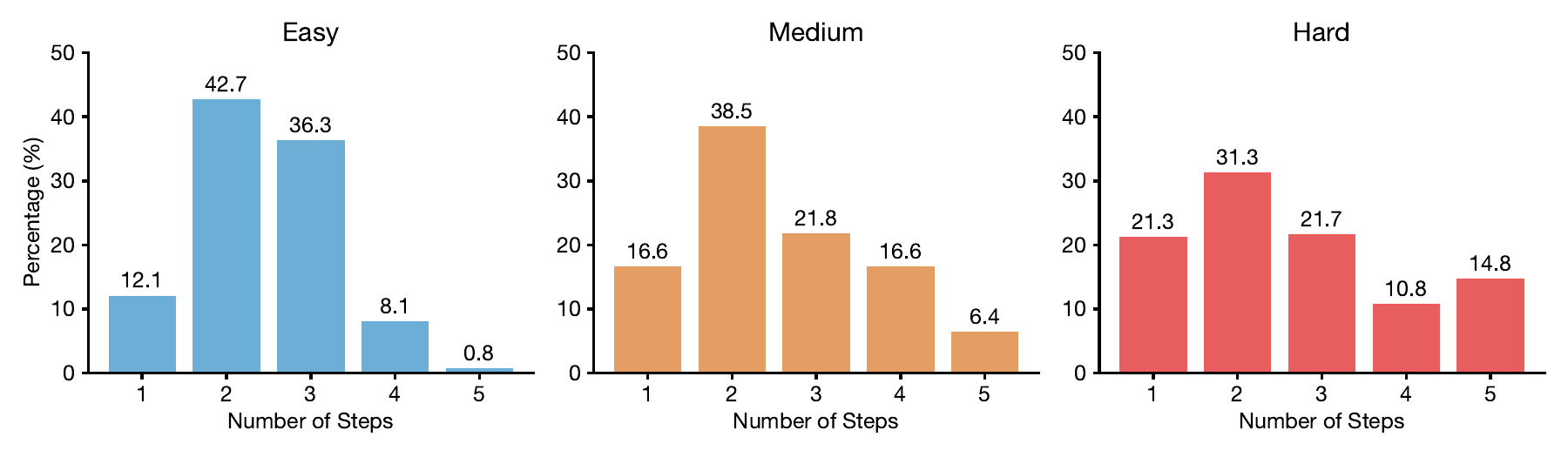}
    \caption{
    \textbf{Distribution of the number of PRM steps defined by FunPRM.} Results are obtained from O4-mini (High)-generated code on LiveCodeBench across easy, medium, and hard categories.
    }
    \label{fig:dist}
\end{figure*}

In this section, we examine whether FunPRM generalizes to other coding datasets when applied directly to HumanEval+ and MBPP+ without additional training. As shown in Figure~\ref{fig:domain-gen}, FunPRM continues to improve the performance of the base LLM through test-time scaling on these benchmarks. This result suggests that FunPRM learns to assign meaningful rewards to coding tasks with diverse formats, demonstrating an ability to assess Python code quality rather than overfitting to a specific benchmark or platform, and highlighting its strong domain generalization capability.

\begin{figure}[h!]
    \centering
    \includegraphics[width=0.8\linewidth]{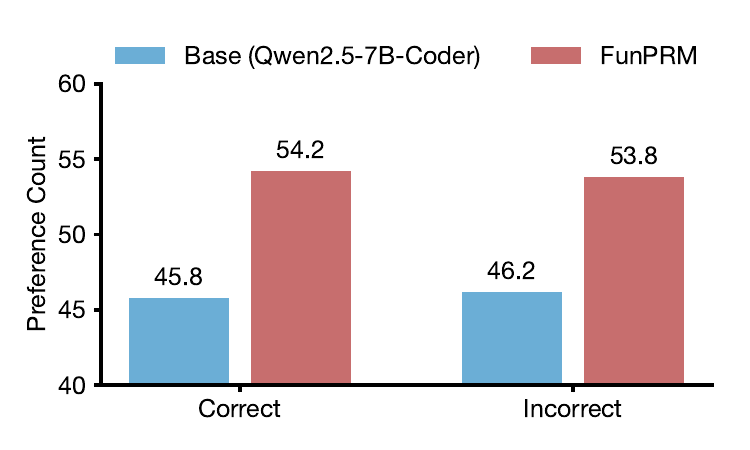}
    \caption{
    \textbf{Human evaluation of code quality for FunPRM and the base LLM (Qwen2.5-7B-Coder).}
    The results report the number of coder preferences for problems where both methods produce correct solutions (left) and incorrect solutions (right).
    }
    \label{fig:human-eval}
\end{figure}

\subsection{Human Evaluation}

A key advantage of FunPRM is its modular coding style induced by the Chain-of-Function prompting strategy. To evaluate this property, we conduct a human study comparing the code quality generated by FunPRM and the base LLM. We randomly sample 100 problems from LiveCodeBench for which both FunPRM and the baseline produce correct solutions, and 100 problems for which both produce incorrect solutions. Three graduate-level computer science researchers independently indicate their preference between the two generations. For correct solutions, evaluators focus on code readability, while for incorrect solutions they assess reusability and ease of fixing. As shown in Figure~\ref{fig:human-eval}, FunPRM is consistently preferred over the base model in both settings, highlighting its superior code quality from a human perspective. We show more examples of FunPRM generated code in Appendix~\ref{appen:case}.

\subsection{Ablation Study}

In this section, we conduct an ablation study to examine the contributions of the Chain-of-Function (CoF) prompting strategy and the meta reward correction (MRC) mechanism in FunPRM. We compare the full model against two ablation variants: (1) FunPRM without Chain-of-Function (CoF) prompting, which effectively degenerates into a generative outcome reward model~\citep{zhang2025generative}, and (2) FunPRM without meta reward correction (MRC), which is trained on CoF trajectories using standard Monte Carlo-sampled labels. As shown in Table~\ref{tab:abl}, FunPRM consistently outperforms both ablation variants, demonstrating the effectiveness of both the CoF formulation and the meta-learning-based reward correction framework.

\begin{table}[t]
\centering
\caption{\textbf{Ablation results on LiveCodeBench (2025-02-01 -- present).} Performance comparison between FunPRM and its ablation variants.}
\setlength{\tabcolsep}{5pt}
\begin{tabular}{lcccc}
\toprule
 & Easy & Medium & Hard & Overall \\
 &(31)&(39)&(61)&(131)\\
\midrule
FunPRM w/o CoF & \bf 100 & 89.7 & \bf 63.9 & 80.1 \\
FunPRM w/o MRC  & \bf 100 & \bf 92.3 & 62.3 & 80.2 \\
\bf FunPRM    & \bf 100 & \bf 92.3 & \bf 63.9 & \bf 80.9 \\
\bottomrule
\end{tabular}
\label{tab:abl}
\end{table}

\subsection{Distribution of PRM Steps}

Figure~\ref{fig:dist} shows the distribution of the number of PRM steps across different difficulty levels. Specifically, the code is generated by O4-mini (High) on LiveCodeBench for easy, medium, and hard problems. We observe that as problem difficulty increases, the number of steps (i.e., generated functions) generally increases, indicating that harder problems typically require longer chains of logic to solve. At the same time, despite this shift in the distribution, some hard problems are solved using only a single function. This observation highlights that coding task complexity can remain high even when the solution consists of only a few lines of code. 

\section{Conclusion}
In this work, we introduce FunPRM, a Process Reward Model for LLM-based code generation. FunPRM treats functions as intermediate reasoning steps via a novel Chain-of-Function prompting scheme, enabling effective step decomposition for PRM training and inference in code generation tasks. We further propose a meta-learning-based reward correction framework that leverages clean unit-test-evaluated final-solution rewards to purify noisy Monte Carlo-estimated partial-solution rewards.
Extensive experiments across LiveCodeBench and BigCodeBench demonstrate that FunPRM consistently outperforms existing test-time scaling baselines across multiple base LLMs. When combined with a strong base model, FunPRM achieves state-of-the-art performance on LiveCodeBench. Together, these results highlight FunPRM as a practical and effective approach for improving test-time scaling in LLM-based code generation.

\section{Impact Statements}

In this work, we introduce FunPRM, a Process Reward Model for LLM-based code generation that can be used to better guide code generation at test time. This direction can improve the reliability and usefulness of code generation systems, potentially increasing developer productivity and lowering barriers to programming by enabling faster prototyping and iteration. However, LLM-generated code may still be incorrect or unsafe, and using it without careful inspection can lead to serious harms such as security vulnerabilities, data leakage, or unintended system behavior. These risks may be exacerbated by automation bias and rapid iteration workflows. Responsible deployment should therefore incorporate safeguards such as code review, automated testing, and sandboxed execution that restrict system access before running generated programs, especially in high-stakes or sensitive environments.

\bibliography{example_paper}
\bibliographystyle{icml2026}

\newpage
\appendix
\onecolumn

\section{Approximation of Meta-Gradient}
\label{appen:meta-grad}

Here, we detail an efficient approximation of the meta-gradient in Eq.~\ref{eq:meta}.
For notational convenience, we denote the inner loss on the noisy partial-solution data in Eq.~\ref{eq:lower} as
$
\mathcal{L}_{n}(\phi,\theta)\!=\!\mathcal{L}\!\left(f_\phi(S_n),g_\theta(\widehat{R})\right),
$
and the meta loss on the clean final-solution data in Eq.~\ref{eq:meta} as
$
\mathcal{L}_{m}(\theta)\!=\!\mathcal{L}\!\left(f_{\hat{\phi}}(S_m),R\right).
$
With the one-step inner update $\hat{\phi}=\phi-\eta\nabla_\phi \mathcal{L}_{n}(\phi,\theta)$, the meta loss $\mathcal{L}_{m}(\theta)$ depends on $\theta$ implicitly through $\hat{\phi}$. Differentiating w.r.t.\ $\theta$ yields
\[
\nabla_\theta \mathcal{L}_{m}
=
-\eta\,
\big(\nabla_{\hat{\phi}}\mathcal{L}_{m}\big)^\top
\nabla^2_{\theta,\phi}\mathcal{L}_{n}(\phi,\theta),
\]
which contains an expensive mixed Hessian--vector product. Following DARTS~\citep{liu2018darts}, we avoid explicit Hessian computation via finite differences. Let
$
v=\nabla_{\hat{\phi}}\mathcal{L}_{m}
$
and define perturbed parameters
$
\phi^\pm=\phi\pm\alpha v
$
for a small $\alpha>0$. Then the required product is approximated by
\[
\nabla^2_{\theta,\phi}\mathcal{L}_{n}(\phi,\theta)\,v
\;\approx\;
\frac{\nabla_\theta \mathcal{L}_{n}(\phi^+,\theta)-\nabla_\theta \mathcal{L}_{n}(\phi^-,\theta)}{2\alpha}.
\]
Substituting back gives the practical meta-gradient estimator
\[
\nabla_\theta \mathcal{L}_{m}
\;\approx\;
-\eta\,
\frac{\nabla_\theta \mathcal{L}_{n}(\phi^+,\theta)-\nabla_\theta \mathcal{L}_{n}(\phi^-,\theta)}{2\alpha},
\]
which requires only two backward passes through $\mathcal{L}_{n}$ at $\phi^+$ and $\phi^-$ and avoids forming second-order derivatives explicitly. This approximation has been widely adopted in bi-level optimization~\citep{qin2025bidora} and multi-level optimization~\citep{zhang2025tapweight} based machine learning methods.

\section{Datasets}
\label{appen:data}
\paragraph{LiveCodeBench}
LiveCodeBench (LCB) is a comprehensive and contamination-free benchmark for code generation with large language models~\citep{jain2025livecodebench}. It continuously collects newly released programming problems from competitive programming platforms such as LeetCode, AtCoder, and Codeforces. Each problem is associated with a publication timestamp and categorized into three difficulty levels: easy, medium, and hard. 
In our experiments, we use 601 problems published before 2024-08-01 as the training split for FunPRM and 454 problems published after 2024-08-01 as the primary test set. For the state-of-the-art leaderboard comparison with O4-mini (High) as the base LLM, we additionally report results on a smaller evaluation set of 131 problems published after 2025-02-01, which is a subset of the primary test set, due to computational cost constraints. LiveCodeBench has released multiple dataset versions, and we use the latest version (v6) in this study.

\paragraph{BigCodeBench}
BigCodeBench (BCB) is a benchmark designed to evaluate LLMs on practical and challenging real-world coding tasks, in contrast to the more isolated programming exercises in earlier benchmarks~\citep{zhuo2025bigcodebench}. BCB requires LLMs to invoke multiple function calls as tools drawn from 139 libraries spanning seven domains. The benchmark contains 1{,}140 problems, which are categorized into two difficulty levels: easy and hard\footnote{\url{https://huggingface.co/blog/terryyz/bigcodebench-hard}}.
For BCB, we use 800 problems as training data for FunPRM and the remaining 340 problems as the test split, consisting of 292 easy problems and 48 hard problems.
\paragraph{HumanEval and MBPP}
HumanEval is one of the earliest coding benchmarks and consists of 164 programming problems, each specified by a function signature, a docstring, and a set of unit tests~\citep{Chen2021EvaluatingLL}. The task for the language model is to complete the function given its signature and docstring. MBPP contains 974 crowd-sourced Python programming problems designed to be solvable by entry-level programmers, among which 427 problems are hand-verified~\citep{austin2021program}. Each problem includes a task description, a reference solution, and three automated test cases. In this work, we evaluate on the enhanced EvalPlus versions of these benchmarks, commonly referred to as HumanEval+ and MBPP+~\citep{liu2023is}. 
Compared to the training benchmarks, LiveCodeBench and BigCodeBench, HumanEval and MBPP are easier and typically involve shorter solutions with a single function, representing a modest domain shift. We do not perform any additional training on these datasets; instead, we directly apply FunPRM trained on LiveCodeBench and BigCodeBench to evaluate its domain generalization capability.

\section{Generative Reward Model}
\label{appen:gprm}

Generative reward models use pretrained large language models as backbones and leverage their next-token prediction probabilities to compute reward signals~\citep{zhang2025generative}. Typically, such models employ carefully designed prompts that instruct the LLM to verify the correctness of a candidate solution and output designated tokens, whose generative probabilities are then interpreted as reward scores. FunPRM adopts a generative reward model as its backbone, using the prompt shown in Figure~\ref{fig:reward-model-prompt}.

Given the model outputs, we compute the PRM score using the generative probabilities of the designated tokens, specifically
\[
r = \frac{p^{+}}{p^{+} + p^{-}},
\]
where $p^{+}$ and $p^{-}$ denote the probabilities of the positive and negative verification tokens, respectively.

\begin{figure}[t]
  \centering
  \begin{minipage}{\linewidth}
    \begin{datacollection}
    
\textbf{System Prompt:} You are an advanced AI assistant designed to serve as a process
supervision model for coding problems. The programming problem is described in
\texttt{<problem>...</problem>}, and the solution to be evaluated (which may be partial or
complete) is provided in \texttt{<code>...</code>}. Your role is to assess whether the code
and reasoning expressed so far are correct with respect to the requirements stated in
\texttt{<problem>...</problem>}.

For each step, respond with \texttt{+} if you believe the solution is correct up to and
including the given \texttt{<code>...</code>}, and respond with \texttt{-} if you detect any
issues, errors, or incorrect logic up to this point. Only respond with \texttt{+} or
\texttt{-}. Do not provide any additional explanations, comments, or justifications. Your
task is strictly to verify the correctness of the solution prefix provided so far.
    \end{datacollection}
  \end{minipage}
  \caption{\textbf{System prompt used for the generative process reward model.} The model is instructed to evaluate partial or complete code solutions, and to output a binary correctness signal (\texttt{+}/\texttt{-}) indicating whether the solution prefix satisfies the problem requirements up to the current step.}
  \label{fig:reward-model-prompt}
\end{figure}

\section{Training Settings}
\label{appen:hyperparam}

During the training of FunPRM, we use O4-mini-low~\citep{openai_o3_o4mini_system_card}, Qwen3-Coder-30B-A3B~\citep{Yang2025Qwen3TR}, and DeepSeek-Coder~\citep{Guo2024DeepSeekCoderWT} to generate Chain-of-Function (CoF) trajectories for PRM training. We employ multiple models from different organizations with varying capabilities to ensure diversity in coding styles within the training data, thereby improving the generalizability of the PRM. We then use Qwen3-Coder-30B-A3B~\citep{Yang2025Qwen3TR} to generate Monte Carlo (MC)-sampled rewards for partial solutions, following the strategies used in prior work~\citep{wang-etal-2024-math,wang-etal-2024-multi-step}, which are subsequently used for training the process reward model. These MC-estimated rewards are used only as initialization for the optimizable reward parameters, which are later refined under the FunPRM framework. 

We train FunPRM using the Adam optimizer~\citep{Kingma2014AdamAM} with a learning rate of $10^{-4}$, a meta-learning rate of $10^{-3}$, a weight decay of $10^{-3}$, and a batch size of 16. Training is conducted for 20{,}000 iterations on two NVIDIA A100 GPUs. We train the PRM using bfloat16 (bf16) precision. LoRA is applied with rank $r=8$, scaling factor $\alpha=16$, and dropout rate $0.05$, targeting the query, key, value, and output projection layers of the transformer~\citep{hu2022lora}. 

\section{Baselines}
\label{appen:baseline}

\paragraph{Test-time Scaling Baselines}
Self-Certainty is proposed to address the limitation that self-consistency cannot be applied to open-ended generation tasks~\citep{kang2025scalable}. It leverages the inherent probability distrfibution of LLMs and selects the candidate solution with the highest divergence between the predicted token distribution and a uniform distribution. A distribution that diverges significantly from uniform indicates a more peaked—and thus more certain—prediction.
Outcome Reward Models (ORMs) assign a single reward to the entire generated solution rather than step-wise rewards~\citep{Cobbe2021TrainingVT,Uesato2022SolvingMW}. Concretely, we implement an ORM using Qwen-2.5-Coder-7B as the backbone and replace its final language-model head with a classification head to output a scalar reward score. This model is trained on the same training problems used for FunPRM, but only on final solutions, without using partial-solution data.
Skywork-PRM-7B is a process reward model that can be applied to both mathematical reasoning and coding tasks~\citep{He2025SkyworkOR}. It is fine-tuned from a Qwen-2.5-7B model, and the PRM is open-sourced\footnote{\url{https://huggingface.co/Skywork/Skywork-o1-Open-PRM-Qwen-2.5-7B}}.

\paragraph{LLM Baselines}
In addition, we report results for strong proprietary and open-weight LLMs on LiveCodeBench without test-time scaling. The baselines reported in Table~\ref{tab:leaderboard} include O4-mini (High)~\citep{openai_o3_o4mini_system_card}, a  reasoning model optimized for strong coding and STEM performance; Gemini-2.5~\citep{Comanici2025Gemini2P}, a flagship multimodal model family designed for strong general reasoning and code generation; DeepSeek-R1~\citep{guo2025deepseekr1}, an open-weight reasoning model trained for competitive step-by-step reasoning and coding; as well as O3 (High)~\citep{openai_o3_o4mini_system_card}, a higher-capability variant of OpenAI's O3 reasoning series intended for more challenging reasoning tasks. Additional LLM baselines shown in Figure~\ref{fig:leaderboard} include EXAONE-4.0~\citep{exaone-4.0}, a recent open-weight LLM emphasizing reasoning and multilingual capabilities, and O3-mini~\citep{openai_o3_o4mini_system_card}, a smaller reasoning-focused model designed for improved efficiency.

\section{Case Study}
\label{appen:case}
In this section, we present qualitative examples comparing code generated by a base LLM and by FunPRM to provide an intuitive understanding of Chain-of-Function-style generation. We first consider a LeetCode-style coding problem shown in Figure~\ref{fig:coding-problem-leetcode}, which requires repeatedly transforming a string and querying a specific character position. The corresponding solutions generated by the base LLM and FunPRM are shown in Figure~\ref{fig:code-leetcode}. While the baseline solution implements all logic within a single function, the FunPRM-generated solution decomposes the computation into helper functions with clear docstrings, explicitly separating string transformation from the main control flow. As required by the LeetCode format, both solutions are wrapped within a \texttt{Solution} class.

We further examine an AtCoder-style coding problem in Figure~\ref{fig:coding-problem-atcoder}, which asks whether a target sequence appears at least twice as distinct subsequences within a given array under large input constraints. The base LLM solution shown in Figure~\ref{fig:code-atcoder-baseline} implements the logic in a monolithic \texttt{main} function, directly computing the earliest and latest matching positions. In contrast, the FunPRM-generated solution in Figure~\ref{fig:code-atcoder-funprm} modularizes the algorithm into dedicated helper functions for computing earliest and latest subsequence matches, each accompanied by descriptive docstrings. 

\clearpage

\begin{figure}[h!]
  \centering
  \begin{minipage}{\linewidth}
        \begin{datacollection}
        
\textbf{Problem.} \\

Alice and Bob are playing a game. Initially, Alice has a string \texttt{word = "a"}.
You are given a positive integer $k$.

\medskip
Bob asks Alice to repeat the following operation forever:

\medskip
Generate a new string by changing each character in \texttt{word} to its next character in the English alphabet, and append it to the original \texttt{word}.

\medskip
For example, performing the operation on \texttt{"c"} generates \texttt{"cd"}, and performing the operation on \texttt{"zb"} generates \texttt{"zbac"}.
Return the value of the $k$-th character in \texttt{word}, after enough operations have been done for \texttt{word} to have at least $k$ characters.
Note that the character \texttt{"z"} can be changed to \texttt{"a"} in the operation.
    \end{datacollection}
  \end{minipage}
  \caption{\textbf{Example of a LeetCode-style coding problem.} The task involves repeatedly transforming a string by appending a character-shifted copy of itself and requires determining the $k$-th character in the resulting string after sufficient iterations.}
  \label{fig:coding-problem-leetcode}
\end{figure}

\begin{figure}[h!]
  \centering
  \begin{minipage}{\linewidth}
    
    \begin{datacollection}{}
{\bfseries Problem.}

\medskip
There is a sequence $A = (A_1,\dots,A_N)$. Determine whether there are at least two subsequences of $A$
that match the sequence $B = (B_1,\dots,B_M)$. Two subsequences are distinguished if they are taken
from different positions, even if they coincide as sequences.\\

\textbf{Subsequence.} 

\medskip
A subsequence of $A$ is a sequence obtained by removing zero or more elements from $A$ and leaving
the remaining elements in their original order. \\

\textbf{Input.}
\medskip

The input is given from Standard Input in the following format:
\[
\begin{array}{l}
N\; M\\
A_1\; A_2\; \ldots\; A_N\\
B_1\; B_2\; \ldots\; B_M
\end{array}
\]
\\

\textbf{Output.} 
\medskip

If there are at least two subsequences of $A$ that match $B$, print \texttt{Yes}. Otherwise, print \texttt{No}.
\\

\textbf{Constraints.}
\medskip

\[
\begin{aligned}
&1 \leq M \leq N \leq 2\times 10^5\\
&1 \leq A_i \leq 10^9\\
&1 \leq B_i \leq 10^9
\end{aligned}
\]
All input values are integers.
    \end{datacollection}
  \end{minipage}
  \caption{\textbf{Example of an AtCoder-style coding problem.} 
    The task asks whether a target sequence appears at least twice as distinct subsequences within a given array, requiring efficient reasoning over subsequence matching under large input constraints.}
  \label{fig:coding-problem-atcoder}
\end{figure}

\begin{figure}[t!]
  \centering
  \begin{minipage}{\linewidth}
    \small
    \begin{datacollection}{}
{\bfseries \normalsize Baseline Code Solution.}\\
{\ttfamily
\obeylines
class Solution:
\codeindent def kthCharacter(self, k: int) -> str:
\codeindenttwo \# Start with the initial word
\codeindenttwo word = "a \\
\codeindenttwo \# Keep expanding until we have at least k characters
\codeindenttwo while len(word) < k:
\codeindentthree \# Build the transformed string
\codeindentthree transformed = [] \\
\codeindentthree for c in word:
\codeindentthree\codeindent \# Compute next character in a cyclic manner
\codeindentthree\codeindent next\_char = chr((ord(c) - ord('a') + 1) \% 26 + ord('a'))
\codeindentthree\codeindent transformed.append(next\_char)
\codeindentthree \# Append to the original word
\codeindentthree word += "".join(transformed) \\
\codeindenttwo \# Return the k-th character (1-based index)
\codeindenttwo return word[k-1]\\
}
{\bfseries \normalsize FunPRM Generated Code Solution.}\\
{\ttfamily
\setlength{\parskip}{0pt}\setlength{\parindent}{0pt}%
\obeylines
class Solution:
\codeindent def kthCharacter(self, k: int) -> str:
\codeindenttwo """
\codeindenttwo Simulate iterative string doubling by appending next-letter transformation (\_next\_transform())
\codeindenttwo until length >= k. Return the k-th character (1-indexed).
\codeindenttwo """
\codeindenttwo word = "a"
\codeindenttwo while len(word) < k:
\codeindentthree word += self.\_next\_transform(word)
\codeindenttwo return word[k - 1]\\ 

\codeindent def \_next\_transform(self, s: str) -> str:
\codeindenttwo """
\codeindenttwo Return string formed by shifting each character of s to its next letter, wrapping 'z' to 'a'.
\codeindenttwo ord('a') = 97, ord('z') = 122
\codeindenttwo """ 
\codeindenttwo res = []
\codeindenttwo for ch in s:
\codeindentthree code = ord(ch) + 1
\codeindentthree if code > ord('z'):
\codeindentthree\codeindent code = ord('a')
\codeindentthree res.append(chr(code))
\codeindenttwo return "".join(res)
}
\end{datacollection}
  \end{minipage}
    \caption{\textbf{Comparison of baseline and FunPRM-generated solutions for a LeetCode-style problem.} The baseline solution implements the string expansion logic inline within a single function, while the FunPRM-generated solution decomposes the logic into functions.}
  \label{fig:code-leetcode}
\end{figure}

\begin{figure}[t]
  \centering
  \begin{minipage}{\linewidth}
    \begin{datacollection}{}
{\bfseries Baseline Code Solution.}\\
{\ttfamily
\setlength{\parskip}{0pt}\setlength{\parindent}{0pt}%
\obeylines
import sys
import threading

def main():
\codeindent input = sys.stdin.readline
\codeindent N, M = map(int, input().split())
\codeindent A = list(map(int, input().split()))
\codeindent B = list(map(int, input().split())) \\

\codeindent \# Compute the earliest positions L[k] in A matching B[0..k]
\codeindent L = [0] * M
\codeindent ptr = 0
\codeindent for i, a in enumerate(A):
\codeindenttwo if ptr < M and a == B[ptr]:
\codeindentthree L[ptr] = i
\codeindentthree ptr += 1
\codeindentthree if ptr == M:
\codeindentthree\codeindent break
\codeindent if ptr < M:
\codeindenttwo \# B cannot be matched even once
\codeindenttwo print("No")
\codeindenttwo return \\

\codeindent \# Compute the latest positions R[k] in A matching B[k..M-1]
\codeindent R = [0] * M
\codeindent ptr = M - 1
\codeindent for i in range(N - 1, -1, -1):
\codeindenttwo if ptr >= 0 and A[i] == B[ptr]:
\codeindentthree R[ptr] = i
\codeindentthree ptr -= 1
\codeindentthree if ptr < 0:
\codeindentthree\codeindent break
\codeindent if ptr >= 0:
\codeindenttwo \# Safe guard if backward scan fails
\codeindenttwo print("No")
\codeindenttwo return \\

\codeindent \# If any L[k] < R[k], two distinct subsequences exist
\codeindent for k in range(M):
\codeindenttwo if L[k] < R[k]:
\codeindentthree print("Yes")
\codeindentthree return \\

\codeindent \# Otherwise the matching positions are unique
\codeindent print("No")
}
    \end{datacollection}
  \end{minipage}
    \caption{\textbf{Baseline solution for an AtCoder-style coding problem.} 
    The code checks whether a target sequence appears at least twice as distinct subsequences by computing the earliest and latest matching positions of each element and verifying whether multiple valid matchings exist.}
  \label{fig:code-atcoder-baseline}
\end{figure}

\begin{figure}[t]
  \centering
  \begin{minipage}{\linewidth}
    \small
    \begin{datacollection}{}
{\bfseries \normalsize FunPRM Generated Code Solution.}\\
{\ttfamily\small
\setlength{\parskip}{0pt}\setlength{\parindent}{0pt}%
\obeylines
def main():
\codeindent """
\codeindent Determines whether there are at least two distinct subsequences of A that equal B.
\codeindent The approach computes the earliest and latest possible index matches of B as a subsequence of A. 
\codeindent If B cannot be matched at all, the answer is "No". 
\codeindent If any position in the earliest match is strictly earlier than the corresponding position in the latest match, then multiple distinct subsequences exist.
\codeindent """
\codeindent input = sys.stdin.readline
\codeindent N, M = map(int, input().split())
\codeindent A = list(map(int, input().split()))
\codeindent B = list(map(int, input().split()))

\codeindent L = find\_earliest\_positions(A, B)
\codeindent if L is None:
\codeindenttwo print("No")
\codeindenttwo return

\codeindent R = find\_latest\_positions(A, B)
\codeindent for k in range(M):
\codeindenttwo if L[k] < R[k]:
\codeindentthree print("Yes")
\codeindentthree return

\codeindent print("No")\\

def find\_earliest\_positions(A, B):
\codeindent """
\codeindent Greedily matches B as a subsequence of A using the earliest possible indices.
\codeindent Returns a list L where L[k] is the index in A matching B[k], or None if B
\codeindent is not a subsequence of A.
\codeindent """
\codeindent N, M = len(A), len(B)
\codeindent L = [0] * M
\codeindent j = 0
\codeindent for k in range(M):
\codeindenttwo while j < N and A[j] != B[k]:
\codeindentthree j += 1
\codeindenttwo if j == N:
\codeindentthree return None
\codeindenttwo L[k] = j
\codeindenttwo j += 1
\codeindent return L \\

def find\_latest\_positions(A, B):
\codeindent """
\codeindent Greedily matches B as a subsequence of A using the latest possible indices.
\codeindent Returns a list R where R[k] is the index in A matching B[k]. Assumes that
\codeindent B is a subsequence of A.
\codeindent """
\codeindent N, M = len(A), len(B)
\codeindent R = [0] * M
\codeindent j = N - 1
\codeindent for k in range(M - 1, -1, -1):
\codeindenttwo while j >= 0 and A[j] != B[k]:
\codeindentthree j -= 1
\codeindenttwo R[k] = j
\codeindenttwo j -= 1
\codeindent return R
}
    \end{datacollection}
  \end{minipage}
  \caption{\textbf{FunPRM-generated solution for an AtCoder-style coding problem.}
    The solution decomposes the logic into functions , separately computing the earliest and latest subsequence matches to determine whether multiple valid subsequences exist.}
  \label{fig:code-atcoder-funprm}
\end{figure}


\end{document}